\documentclass{ecai} 

\usepackage{latexsym}
\usepackage{amssymb}
\usepackage{amsmath}
\usepackage{amsthm}
\usepackage{booktabs}
\usepackage{enumitem}
\usepackage{graphicx}
\usepackage{color}

\usepackage[ruled, vlined, linesnumbered]{algorithm2e}
\usepackage{caption}
\usepackage{subcaption}

\newcommand{\BibTeX}{B\kern-.05em{\sc i\kern-.025em b}\kern-.08em\TeX}

\begin{document}

\begin{frontmatter}

\title{Distilling Reinforcement Learning into Single-Batch Datasets}

\author{\fnms{Connor}~\snm{Wilhelm}}
\author{\fnms{Dan}~\snm{Ventura}}

\address{Brigham Young University}

\begin{abstract}
Dataset distillation compresses a large dataset into a small synthetic dataset such that learning on the synthetic dataset approximates learning on the original. Training on the distilled dataset can be performed in as little as one step of gradient descent. We demonstrate that distillation is generalizable to different tasks by distilling reinforcement learning environments into one-batch supervised learning datasets. This demonstrates not only distillation's ability to compress a reinforcement learning task but also its ability to transform one learning modality (reinforcement learning) into another (supervised learning). We present a novel extension of proximal policy optimization for meta-learning and use it in distillation of a multi-dimensional extension of the classic cart-pole problem, all MuJoCo environments, and several Atari games. We demonstrate distillation's ability to compress complex RL environments into one-step supervised learning, explore RL distillation's generalizability across learner architectures, and demonstrate distilling an environment into the smallest-possible synthetic dataset.
\end{abstract}

\end{frontmatter}

\section{Introduction}

As deep learning increases in popularity and models are trained on larger tasks, training costs impose a greater cost to the deep learning community and the planet. High training costs provide a high barrier to entry to many, and can make hyperparameter searches such as neural architecture search prohibitively expensive. Training's high energy usage creates high economic and ecological costs. As deep learning advances, methods to reduce training costs become more necessary. Reinforcement learning in particular suffers from high training costs from repeated simulations of the environment, inefficient experience gathering, reliance on auxiliary networks, and many potential failure states. We propose an extension of dataset distillation that can reduce reinforcement learning costs to near-negligible levels: a single step of gradient descent, while avoiding expensive environment simulations.

Dataset distillation\footnote{In this work, we utilize ``distillation" to refer only to dataset distillation and its generalization, task distillation; not to knowledge/policy distillation.} is a technique in which a large dataset is compressed into a synthetic dataset, learnable in as few as a single step of gradient descent \citep{dataset_distillation}. The performance of a model when trained on the synthetic dataset should approximate the performance of that model when trained on the original dataset. By compressing the original dataset, distillation greatly reduces the cost of training models; in this work, we demonstrate one-step learning made possible with distillation. The compression provided by distillation can be used for low-resource training, interpretability, and data anonymization \citep{data_anonymity}. The reduction of training costs resulting from this compression can accelerate searches such as hyperparameter searches, neural architecture searches, and high-performance initialization searches, and can produce inexpensive ensembles.

We propose a generalization of dataset distillation, which we call \textit{task distillation}. Using techniques similar to dataset distillation, we posit that any learning task can be compressed into a synthetic task. Task distillation allows for transmodal distillations, reducing a more complex task, such as a reinforcement learning environment, to a simpler task, such as a supervised classification dataset.

In order to advance task distillation beyond supervised-to-supervised distillation, we explore distilling reinforcement learning environments into supervised learning datasets, which we call \textit{RL-to-SL distillation}. We choose to produce synthetic supervised datasets because supervised learning is the bread-and-butter of machine learning, with decades of research going into improving SL. We consider transforming an arbitrary learning task into supervised learning to be an important step to simplify machine learning. We select RL environments to distill because reinforcement learning has the added complexity of requiring exploration of the environment. Exploration can cause a host of difficulties in learning, such as poor policy spaces that an agent cannot escape and a general increase in training complexity. Learning a successful policy often requires a great deal of exploration, which may involve learning on many repetitive observations that hold little useful information. By producing a compressed synthetic dataset, agents can learn an RL task quickly without exploration costs, as exploration has already occurred during distillation. 

The paper begins by exploring distillation on the cart-pole problem \citep{cartpole} as a pedagogical exercise; then demonstrates distillation of more difficult RL environments [MuJoCo \citep{mujoco}, \textit{Centipede} \citep{centipede}, \textit{Ms.\ Pac-Man} \citep{mspacman}, \textit{Pong} \citep{pong}, and \textit{Space Invaders} \citep{spaceinvaders}]; and
makes the following contributions: 
\begin{enumerate}
    \item Proposes a new formulation of RL task distillation using proximal policy optimization.
    \item Proposes an N-dimensional extension of the cart-pole environment to allow for scaling the difficulty of cart-pole.
    \item Demonstrates $k$-shot learning on single-batch datasets distilled from ND cart-pole; demonstrating generalization to unseen initialization distributions architectures.
    \item Empirically validates the theoretical work of \citet{less_than_one_shot} by demonstrating the minimum distillation sizes of environments with different numbers of action classes.
    \item Provides a method that can be used to scale the difficulty of distillation for complex tasks.
    \item  Demonstrates distillation of Atari and MuJoCo environments. 
\end{enumerate}

We provide additional information, including experiment details, ND cart-pole details, and additional figures and results in the appendices. The GitHub repository containing experiment code and distilled datasets are archived on Zenodo \cite{this_code}.

\section{Background}

\subsection{Dataset Distillation}

Dataset distillation was originally proposed by \citet{dataset_distillation}. This method utilizes meta-gradients to train a synthetic dataset, based on how well a predictive model trained on the synthetic data performs on the true dataset. This process is split into two nested learning loops: in inner learning, a newly-initialized predictive model is trained on the synthetic dataset. In outer learning, the trained predictive model is tested against the true dataset, and the loss is backpropagated through the inner-learning process to the synthetic dataset. The synthetic dataset is updated with gradient descent and is used to train a new predictive model in the next iteration.

The goal of dataset distillation is to produce a synthetic dataset $\{X_d, Y_d\}_\theta$ such that predictive models sampled from a given distribution $\lambda_\phi \in \Lambda$ can train on the synthetic dataset to reach high performance on a targeted training task $T_0$. The original formulation defines a fixed synthetic label vector $Y_d$; however, in this work we use the soft labels of \citet{soft_label}, in which $Y_d$ is a parameterized vector of real numbers trained along with the synthetic data instances. The sampled learner is trained on the synthetic dataset using an appropriate loss measure; we utilize mean-squared error between prediction logits and soft labels. This inner optimization must be differentiable, as outer learning requires backpropagating through the inner optimization. The trained model is tested against a sample of $T_0$ with an appropriate loss metric (generally one that would be used in learning on $T_0$ directly) to produce the outer meta-loss. 

It is expected that the distillation's performance is on average less than or equal to the performance of the average model trained directly on the task. This is an expected trade-off for high compression and low training time, as the compressed dataset cannot contain all the information found in the full task. However, closely approaching this performance is vital for distillation's use cases.

\begin{figure}[t]
        \centering
        \includegraphics[width=.5\textwidth]{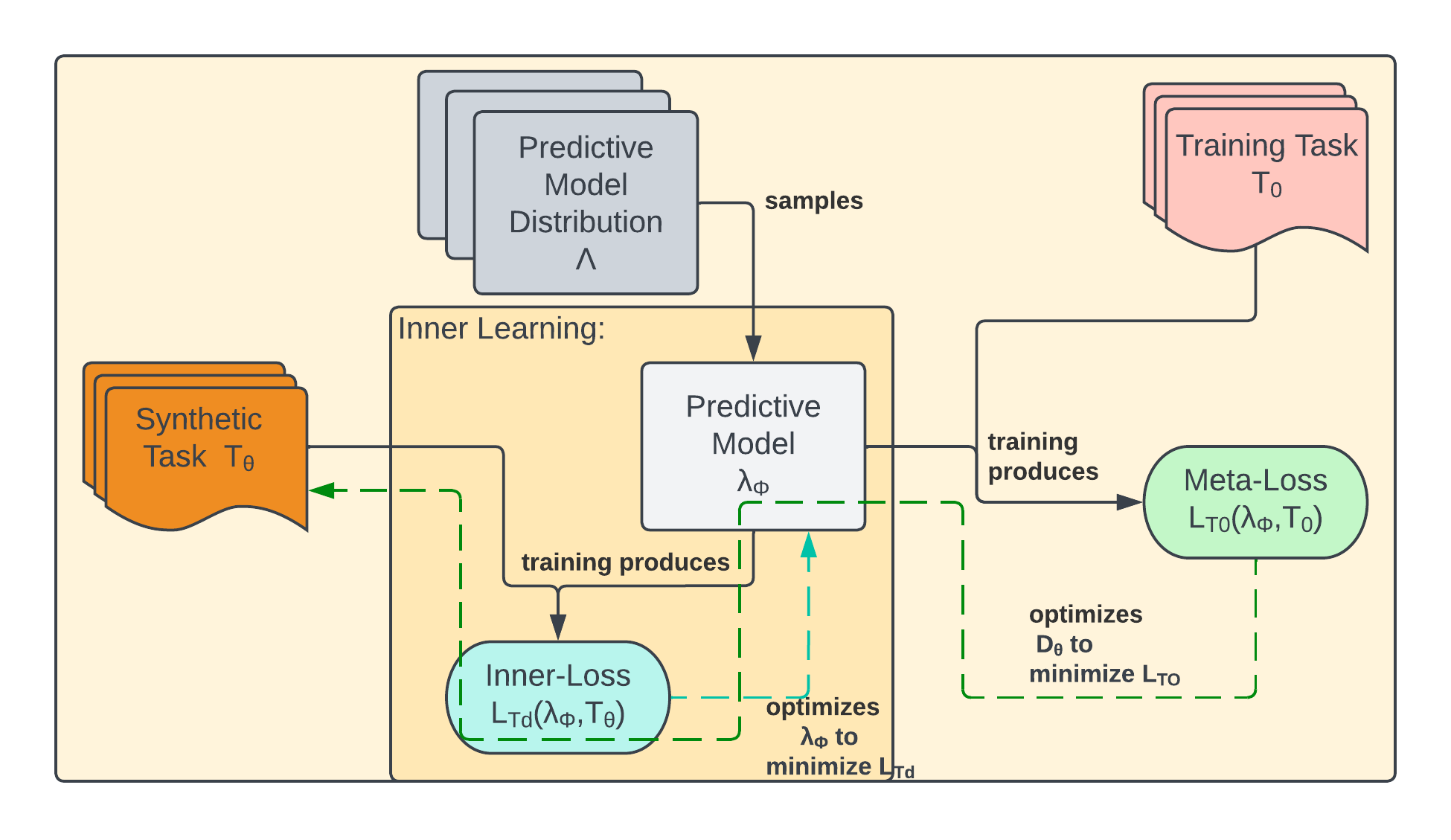}
        \caption{Training process for task-agnostic distillation based on \citet{dataset_distillation}}
        \label{fig:TaskDistill}

        \vspace{20 pt}
\end{figure}

We visualize a task-agnostic version of this formulation of dataset distillation in Figure \ref{fig:TaskDistill}.

\subsection{Related Works in Dataset Distillation}

A similar method to dataset distillation was proposed by \citet{gtn}. Their method uses meta-gradients to train a generator model to produce synthetic datasets, rather than directly learning one dataset. This method, the generative teaching network, was tested on supervised learning tasks and on the simple RL cart-pole environment using A2C. Our experiments are inspired by this work, though we base our method on that of \citet{dataset_distillation} as our preliminary experiments indicated that parameterizing the dataset provided better results.

Other distillation methods exist, such as gradient matching \citep{dataset_condensation}, trajectory matching \citep{dd_exp_traj}, distribution matching \citep{dc_distribution_matching}, differentiable Siamese augmentation \citep{dc_siamese}, and kernel ridge-regression \citep{dd_kernel}. We base our formulation of that of \citet{dataset_distillation} because of its simplicity and its independence from expert models and trajectories, though we posit that our method could be adapted to any of these formulations.

Non-distillation methods for simplifying learning, such as imitation learning, also require expert examples. Unlike distillation, imitation learning trains a single student model per run \cite{imitation_learning}, while distillation produces a dataset that can produce many trained models inexpensively. Thus, imitation learning and distillation have very different use cases.

\citet{dd_orl} distills offline RL environment dataset. Our work distinguishes itself by focusing on online RL learning---sampling data by interacting with the environment (semi-supervised learning) rather than using a presampled offline dataset (supervised learning).

\citet{behaviour_distillation} has similar motivations to our work and also tackles distilling reinforcement learning environments. Their approach uses evolutionary strategies to optimize the dataset, while we utilize meta-gradients calculated with PPO loss. Our work distinguishes itself by focusing on minimum-sized datasets and single-step learning, which is feasible given the direction provided by the meta-gradients.

\subsection{Proximal Policy Optimization}

PPO is a reinforcement learning algorithm that combines two policy gradient methods: actor-critic and trust region optimization. PPO uses two networks: the actor $\pi_\phi$ to learn the optimal policy, and the critic $V_\psi$ to approximate the value function given policy $\pi_\phi$. The actor determines the actions taken at each timestep at training and evaluation, while the critic evaluates the actions taken during training only; it performs no function at evaluation time. PPO actor loss over a batch of a given trajectory of observations is as follows \citep{ppo}:

\begin{equation}
  \begin{split}
    L_{\pi} = - \frac{1}{b} \sum_1^b \min \Bigg[\frac{\pi_\phi^i(a_t | s_t)}{\pi_\phi^0(a_t | s_t)} & A_t, \\
    \text{clip} \left (\frac{\pi_\phi^i(a_t | s_t)}{\pi_\phi^0(a_t | s_t)}, 1-\epsilon, 1+\epsilon \right ) & A_t \Bigg]
  \end{split}
  \label{eq:PPO_pi_loss}
\end{equation}

\noindent where $b$ is the batch size, $\pi_\phi^0$ is the policy network with the parameters used to gather the environment transitions, $\pi_\phi^i$ is the policy network with its current parameters, $a_t$ is the action taken and $s_t$ is the state at time $t$, $A_t$ is the advantage (which we calculate with generalized advantage estimation \cite{gae}), and $\epsilon$ is the trust region width.

PPO requires an auxiliary critic network, trained alongside the actor, to determine $A_t$ for $L_\pi$. The critic is trained to approximate the return $R_t$ expected from timestep $t$, given the agent is at state $s_t$ and is following the policy $\pi_\phi$. Batched critic loss is as follows \citep{ppo}:

\begin{equation}
    L_{V} = \frac{1}{b} \sum_1^b (V_\psi(s_t) - R_t)^2
    \label{eq:PPO_critic_loss}
\end{equation}
\noindent where $b$ is batch size, $V_\psi$ is the critic network, $s_t$ is the state at time $t$, and $R_t$ is the return at time $t$.

\section{RL-to-SL Distillation}

The learner $\lambda$---distillation-trained predictive model---must be capable of performing on both the true task $T_0$ (i.e. act as an RL agent) and the distilled task $T_d$ (i.e. act as a classifier).
In RL-to-SL distillation, $\lambda$ must be capable of both producing a policy $\hat{p}$ for acting in the outer RL environment $E_0$ based on a state observation: $\hat{p} = \lambda(s)$ for all state observations $s$ in $E_0$, and label prediction in the inner synthetic SL task based on a data instance: $\hat{y} = \lambda(x)$, for all $x \in X_d$. Once distillation is complete, a new learner $\lambda \in \Lambda$ trained on $\{X_d, Y_d\}_\theta$ should solve the environment $E_0$, or achieve an acceptable episodic reward on $E_0$ without training on $E_0$.

We define the synthetic data instances and labels to match the dimensionality of the states and actions of the RL environment, respectively. Thus, training the learner, the RL policy network, on the synthetic task is equivalent to learning to classify synthetic states into an action or regressing a state-dependent policy. However, the synthetic data instances are not guaranteed to be in or near the state space. The only constraint on the distilled dataset is how well the learners trained on it perform on $E_0$.

The distillation meta-learning process only distills information pertinent to performing on the task. Thus, when we discuss distilling an environment, we mean that the learning trajectories and policies are distilled, not aspects such as transition and reward functions.

\subsection{Proximal Policy Meta-Optimization} \label{sec:alg_select}

We consider PPO to be a great candidate for the distillation outer-learning objective due to the benefits of its trust-region protections. However, a direct adaptation of dataset distillation using PPO as the outer objective is not sufficient to utilize these trust region protections. PPO's trust region is created by the clip function, which sets the gradient to 0 when the change in policy throughout learning on a given RL trajectory exceeds a threshold set by $\epsilon$. This is because once clipping is applied, infinitesimal changes will not affect the result. This requires that the policy network $\pi_\phi$ be altered throughout training on the RL trajectory. However, only the synthetic dataset is changed by the policy loss during distillation. Thus, we alter dataset distillation such that the agents' initializations do not change throughout training on a given trajectory (i.e. over a single PPO epoch) to ensure that the only changes reflected in the change in policies is due to optimization on PPO loss, rather than factors such as parameter initialization or model architecture.

We propose a new formulation for task distillation using PPO as the outer objective, and introduce it as \textit{Proximal Policy Meta-Optimization} (PPMO) for RL-to-SL distillation. This incorporates the policy trust region protections of PPO.

\begin{algorithm}[t]
  \SetAlgoLined
  \SetKwInOut{Input}{input}
  \SetKwInOut{Output}{output}
  
  \Input{initialized synthetic dataset $\{X_d, Y_d\}_\theta$, learner distribution $\Lambda$, value network $V_\psi$, RL environment $E_0$, number of episodes per iteration $e$, number of policy epochs $n$, RL batch size $b$}
  \Output{learned synthetic dataset $\{X_d, Y_d\}_\theta$ distilled from $E_0$}
  \BlankLine
  \While{$\{X_d,Y_d\}_\theta$ has not converged}{ \label{PPO:metaepoch}
    $\lambda_{\phi_{init}}$ := Sample($\Lambda$)\; \label{PPO:lambda}
    $\pi_\phi^0$, $\nabla_\phi$ := Train($\lambda_{\phi_{init}}$, $\{X_d, Y_d\}_\theta$)\; \label{PPO:innertrain}
    $\tau$ = PerformEpisodes($E_0$, $\pi_\phi^0$, $e$)\; \label{PPO:episodes} 
    $i := 0$\;
    \For{policy epoch from $1$ to $n$}{ \label{PPO:policyepoch}
        \For{$(s, a, r) \in \tau$ of batch size $b$}{ \label{PPO:batches}
            $L_\pi$, $L_V$ := PPOLoss($\pi^i_\phi$, $\pi^0_\phi$, $V_\psi$, $s$, $a$, $r$)\;
            $\nabla\theta$ := BackpropagateWithMetaGradients($L_\pi$, $\theta$, $\nabla_\phi$)\; \label{PPO:ploptim}
            $\nabla\psi$ := Backpropagate($L_V$, $\psi$)\; \label{PPO:vloptim}
            Optimize $\theta$ and $\psi$ w.r.t. $\nabla\theta$ and $\nabla\psi$\;
            $\pi_\phi^{i+1}$, $\nabla_\phi$ := Train($\lambda_{\phi_{init}}$, $\{X_d, Y_d\}_\theta$)\; \label{PPO:retrain}
            $i := i+1$\;
        }
    }
  }
  return $\{X_d, Y_d\}_\theta$\;
  \caption{PPMO for RL-to-SL Distillation($\{X_d, Y_d\}_\theta$, $\Lambda$, $V_\psi$, $E_0$, $e$, $n$, $b$)}
  \label{alg:PPO_distillation}
\end{algorithm}

The process to train the distilled dataset is detailed in Algorithm \ref{alg:PPO_distillation}. Training consists of 3 nested loops: meta-epochs, policy epochs, and batched iterations. At each meta-epoch, a new agent/policy network initialization $\lambda_{\phi_{init}}$ is sampled and trained on the synthetic dataset (lines 2-3). This training uses mean-squared error loss on the model's prediction of synthetic data $X$ against the synthetic label $Y$, and we perform this learning in a single step of gradient descent. The model resulting from the first inner-training process of each meta-epoch, $\pi^0_\phi$, defines the policy executed on the RL environment $E_0$, creating an RL trajectory $\tau$ (line 4). Just like in PPO, this trajectory is used for learning over multiple ``policy epochs". For each batch drawn from the shuffled trajectory, PPO policy and value loss is calculated using the formulas defined in Equations \ref{eq:PPO_pi_loss} and \ref{eq:PPO_critic_loss}, taking in the current parameters of the agent $\pi^i_\phi$, the parameters of the agent used to gather the experiences $\pi^0_\phi$, the value network $V_\psi$, and the batch of experience data (line 8). Our implementation calculates advantage $A_t$ using GAE. The policy loss is our meta-loss: it is backpropagated through the inner-learning process to update the synthetic dataset $\{X_d, Y_d\}_\theta$ (lines 9, 11). By updating the distilled dataset with the policy loss, the distilled dataset is optimized to maximize the reward achieved by the policy produced by inner learning. The critic is updated with standard optimization, i.e. not using meta-learning, as in standard PPO (lines 10-11). A new agent $\pi^{i+1}_\phi$ is trained from the same initialization $\lambda_{\phi_{init}}$ as the previous agents in this meta-epoch (line 12), and is used to calculate loss in the next sampled batch. Once all trajectory batches have been iterated through $n$ times, the next meta-epoch iteration begins by sampling a new agent initialization (line 2). By the end of this process, the synthetic dataset can be used to train any agent initialization in $\Lambda$ to perform on $E_0$.

\section{ND Cart-Pole Experiments}

\begin{figure}
\centering
\begin{subfigure}{.40\textwidth}
  \centering
  \includegraphics[width=\linewidth]{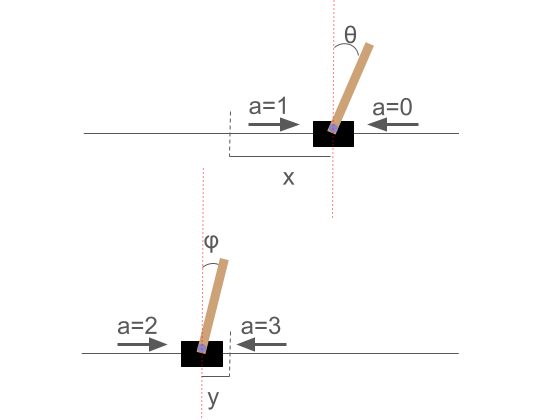}
  \caption{Two perpendicular side views of 2D cart-pole. The solid lines represent the degrees of freedom of the cart.}
  \label{fig:cp2d-sides}
\end{subfigure}
\vspace{12pt}
\begin{subfigure}{.40\textwidth}
  \centering
  \includegraphics[width=\linewidth]{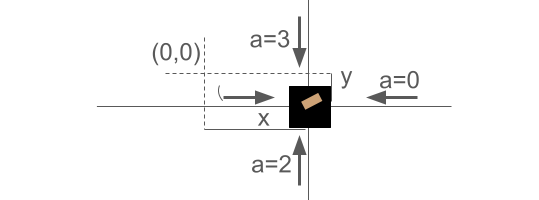}
  \caption{Top-down view of 2D cart-pole. Both dimensions representing the 2 degrees of freedom are visible.}
  \label{fig:cp2d-top}
\end{subfigure}
\vspace{12pt}
\caption{Two views of 2D cart-pole.}
\vspace{20pt}
\label{fig:ndcp}
\end{figure}

To control the difficulty of cart-pole and the size of its state and action spaces, we expand the cart-pole system into multiple dimensions. Classic cart-pole has 1 degree of freedom in which the cart and pole can move. We extend this to $N$ degrees of freedom, though the cart is limited to accelerating along one axis at each timestep, creating an action space of size $2N$. The action space is likewise expanded to $4N$: with one quadruple of cart position, cart velocity, pole angle, and pole angular velocity per degree of freedom. Just as in classic cart-pole, if the pole angle or cart position reach a threshold in any of the $N$ dimensions, the experiment terminates. We provide two visualizations of 2D cart-pole in Figure \ref{fig:ndcp}. We utilize this extension of cart-pole to examine $k$-shot learning and minimum distillation size in RL-to-SL distillation.

Our cart-pole experiments utilize a standard set of hyperparameters unless otherwise stated. We define the learner distribution $\Lambda$ to contain a single architecture, $\alpha_c$, with random initialization. The architecture, initializations, and PPO hyperparameters used in this work are in line with the standard of \citet{shengyi2022the37implementation}, except that we did not use learning rate annealing or value loss clipping.
  
\subsection{\textit{k}-Shot Learning} \label{sec:k-shot}

\begin{table*}[tb]
    \centering
        \caption{k-shot learning on ND cart-pole. Each column represents an agent distribution (initialization and architecture), and each row represents an ND cart-pole distillation, trained only on models sampled from $\Lambda$. Agents are sampled, trained on the distillation, and tested on ND cart-pole, with rewards reported. The distributions are as follows: $\Lambda$ is the agent distribution used in training, using a fixed architecture $\alpha_c$ and orthogonal initialization with a standard deviation $\sigma=0.01$ on the final layer and $\sigma=\sqrt{2}$ on the others. ``$\alpha_c$: Ortho $\sigma$=1'' is like $\Lambda$ but with $\sigma=1$ on all layers. ``$\alpha_c$: Xe'' uses architecture $\alpha_c$ but is initialized using Xe initialization with the same $\sigma$ as $\Lambda$. ``$\alpha_c$: Xe $\sigma$=1'' uses Xe initialization with $\sigma=1$ for all layers. ``Random $h$'' modifies architecture $\alpha_c$ by randomly sampling the hidden layer size from the set $[32, \dots, 256]$ (in $\alpha_c$ the size is fixed at $64$). ``Random $L$'' modifies architecture $\alpha_c$ by randomly sampling the number of hidden layers (number of layers not counting the output layer) from the set $[1,\dots,6]$ (in $\alpha_c$ it is fixed at $2$). Both Random $h$ and Random $L$ used the same initialization as $\Lambda$. The $\Lambda$ column represents base distillation performance, while the others represent generalization to new initializations and architectures.}
    \label{tab:k_shot}\vspace{10 pt}
    \setlength\tabcolsep{6pt}
\begin{tabular}{|c | c | c | c | c | c | c | c |}
  \hline
  Experiment & $\Lambda$ & $\alpha_c$: Ortho $\sigma$=1 & $\alpha_c$: Xe & $\alpha_c$: Xe $\sigma$=1 & Random $h$ & Random $L$ \\
  \hline
  1D 2-Shot & $500 \pm 0.0$ & $238 \pm 200$ & $258 \pm 207$ & $175 \pm 209$ & $500 \pm 0.0$ & $500 \pm 0.0$ \\
  \hline
  1D 512-Shot & $500 \pm 0.0$ & $233 \pm 221$ & $246 \pm 198$ & $137 \pm 178$ & $500 \pm 3.8$ & $500 \pm 0.0$ \\
  \hline
  2D 3-Shot & $365 \pm 31.9$ & $16.3 \pm 11.1$ & $76.8 \pm 67.3$ & $13.4 \pm 6.3$ & $443 \pm 55.7$ & $415 \pm 159$ \\
  \hline
  2D 512-Shot & $350 \pm 37.2$ & $15.9 \pm 10.5$ & $68.6 \pm 50.9$ & $13.9 \pm 7.8$ & $457 \pm 81.0$ & $439 \pm 99.8$ \\
  \hline
  3D 4-Shot & $264 \pm 20.6$ & $12.4 \pm 2.3$ & $38.6 \pm 21.5$ & $10.9 \pm 1.3$ & $379 \pm 93.3$ & $305 \pm 157$ \\
  \hline
  3D 512-Shot & $240 \pm 26.1$ & $12.7 \pm 3.6$ & $41.8 \pm 24.0$ & $11.5 \pm 2.4$ & $401 \pm 132$ & $383 \pm 116$ \\
  \hline
  4D 5-Shot & $127 \pm 7.4$ & $12.5 \pm 2.1$ & $35.0 \pm 12.2$ & $11.0 \pm 1.5$ & $165 \pm 29.8$ & $156 \pm 60.7$ \\
  \hline
  4D 512-Shot & $129 \pm 6.8$ & $12.5 \pm 2.0$ & $39.7 \pm 13.2$ & $11.3 \pm 1.6$ & $203 \pm 55.7$ & $188 \pm 46.1$ \\
  \hline
  5D 6-Shot & $87.5 \pm 4.3$ & $12.8 \pm 1.5$ & $35.7 \pm 9.2$ & $11.5 \pm 1.5$ & $112 \pm 18.3$ & $125 \pm 44.6$ \\
  \hline
  5D 512-Shot & $87.5 \pm 4.0$ & $12.8 \pm 1.8$ & $41.0 \pm 7.1$ & $12.0 \pm 1.5$ & $140 \pm 40.1$ & $159 \pm 30.4$ \\
  \hline
  
\end{tabular}
\end{table*}

Once a synthetic dataset has been fully distilled, it can be used for $k$-shot learning.
New agent models, including those not used in distilling, can be trained on the synthetic task using $k$ learning instances. In our experiments, these $k$ instances are learned as a single batch in one gradient descent step.
As the distilled dataset was trained to teach agents sampled from $\Lambda$, these agents are expected to outperform other agents on the task after $k$-shot learning---outliers within $\Lambda$ and out-of-distribution agents may perform significantly worse on $T_0$ than the average performance of learners from $\Lambda$. However, generalization to unseen architectures is necessary for neural architecture search---a key use-case for distillation.

We verify the success of our distilled datasets by performing $k$-shot learning on randomly sampled agent architectures and initializations in $\Lambda$ on the cart-pole environment (see summary of results in Table \ref{tab:k_shot}). Note that, because the agents in $\Lambda$ used in our experiments share the same architecture but are produced by randomly initializing the parameters, we can assume that no agent used in verification was used in training. We also explore deviations from $\Lambda$ to test $k$-shot learning on out-of-distribution learners. For each distribution tested, we sample 100 model architecture/initialization values, and train each model on the distilled dataset. The model is tested against the RL environment for 100 episodes, and the means and standard deviations of rewards over all the tested models is reported.

Our experiments showed little difference in $k$-shot learning based on the size of $k$: learning on the minimum-sized distillation (see Section \ref{sec:min_dist_size}) generally performed better than 512-shot with agents from $\Lambda$, but performed slightly worse than 512-shot with the other tested distributions. This implies that higher batch sizes are less likely to overfit to $\Lambda$. Our experiments show that $k$-shot learning on $T_d$ appears to be more sensitive to the parameter initialization function (see Ortho and Xe experiments) than to architecture changes (see Random $h$ and $L$ experiments). This strengthens the claim that distillation can be used for neural architecture search. However, these distillations may still be biased to architectures close to those in $\Lambda$; using a wider array of architectures during training would ensure a fairer evaluation of varied neural architectures. We leave a thorough examination of generalization to novel architectures for future work.

\subsection{Minimum Distillation Size} \label{sec:min_dist_size}

Distillation compresses a learning task, transforming it from many-shot learning to finite $k$-shot learning. $k$, the number of distilled instances to produce, is an important hyperparameter in distillation. Our experiments have determined that varying $k$ has only a small effect on the quality of the distillation, past a certain threshold. Increasing $k$ increases the computation costs of the distillation training non-negligibly, as all $k$ instances are updated in each outer learning step and a synthetic dataset with more instances requires more epochs on average to converge. However, the end-learning costs are negligibly increased, so long as $k$ fits into a single batch.

There is, however, a minimum threshold $k_{min}$, such that distillation fails when $k < k_{min}$. This number is determined by the geometry of the label space \citep{less_than_one_shot}. The distilled instances and labels must distinguish the classes or actions from each other. This can be done in fewer instances than the number of classes using soft labels. Using soft labels, the minimum number of instances to distinguish the classes of a label space with $c$ discrete values is:
\begin{equation}
    k_{min} = \lceil c/2 \rceil + 1
    \label{eq:kmin}
\end{equation}
\noindent This allows for what \citet{less_than_one_shot} call ``less than one''-shot learning,\footnote{The term $k$-shot learning is not standardized. In this work, we define $k$ as the number of training instances.} or learning on less than one instance per class.

Equation \ref{eq:kmin} was one of multiple demonstrated by \citet{less_than_one_shot}. We validate this equation experimentally on ND cart-pole and Atari. Distilling to $k \ge k_{min}$ instances results in a successful distillation---the expected reward for $\forall \lambda \in \Lambda$ trained on the distillation is approximately equal to the reward achieved by $\lambda$ after direct RL training. As an example, 2D cart-pole distillation achieves a mean reward of $350$ when $k = 512$ and $365$ mean reward when $k = k_{min} = 3$. When $k < k_{min}$, the distillation is unsuccessful in teaching learners to solve the task, resulting in low average rewards consistent with random performance in cart-pole. 2D cart-pole distilled to $k = k_{min}-1 = 2$ achieves an average of $40.6$ reward.

\begin{figure}[ht]
        \centering
        \includegraphics[width=.5\textwidth]{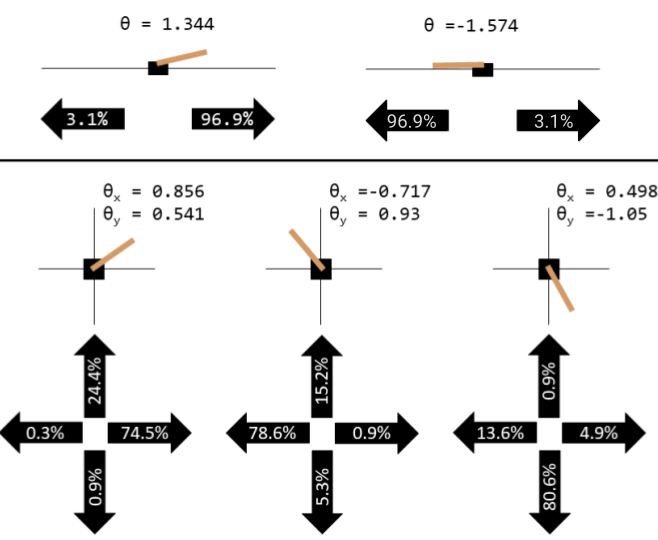}
        \caption{Visualizations of a distillation of 1D (above) and 2D (below) cart-pole with simplified information (only the angle and softmaxed labels). While a full examination of interpretability is outside the scope of this work, this demonstrates that some tasks when distilled provide clear interpretability. In both distillations, the action label demonstrates that the agent should move the cart in the direction in which the pole is leaning. In 2D cart-pole, which was distilled to 3 states despite having 4 actions, the "up" action never has the highest probability mass in any of the labels; instead, the mass is split between two instances. This demonstrates how distillations with fewer than $c$ instances can teach a model to distinguish $c$ classes or actions.}
        \label{fig:distilled_cartpole}
        \vspace{20pt}
\end{figure}

The equation $k_{min} = \lceil c/2 \rceil + 1$ determines how many instances are required to distinguish the classes of the action space, assuming all actions are necessary. In ND cart-pole, all actions are needed to solve the task, and all actions are performed approximately at the same frequency. Thus, the distilled dataset must provide distinction between each action in its soft labels. That is, there must be at least one instance in which the action probability label is high, and one where it is low, relative to the other action probabilities in the label. See Figure \ref{fig:distilled_cartpole} for a visualization of 1D and 2D cart-pole distilled to $k_{min}$ instances.

\section{Atari Environments} \label{sec:atari}

We distill Atari environments to demonstrate RL-to-SL distillation scaling to more complex and difficult reinforcement learning tasks. The Atari environments have a much larger state and action space than cart-pole, as well as more complex reward functions, providing a significant challenge to RL and distillation.

We have chosen a small selection of Atari tasks for brevity. We have avoided selecting tasks that are difficult for a standard PPO agent to learn, as finding tasks that can be learned by distillation but not by direct-task learning is beyond the scope of this work. As Atari tasks are mostly open-ended reward maximization problems, we determine the success of the distillation by comparing the average reward achieved by randomly sampled agent initializations from $\Lambda$ trained on a fully distilled dataset $\{X_d, Y_d\}_\theta$, against the average reward achieved by randomly sampled learners from $\Lambda$ trained directly on the environment using PPO. This is in line with the distillation ratio metric proposed by \citet{soft_label}, which also compares performance after direct-task learning versus distilled-task learning.

For the Atari experiments, we define $\Lambda$ as the distribution of random initializations of a single architecture $\alpha_a$, derived from the standardized PPO implementation of \citet{shengyi2022the37implementation}, as with parameter initialization and all PPO hyperparameters. For the sake of feasibility, we limit the amount of training time for the baseline RL experiments. While it may be possible for the agents to achieve a higher reward given more training time, it is difficult prove RL converged due to large performance plateaus during learning; thus, we consider time limits and apparent convergence as reasonable early-stop heuristics. In order to provide a more fair comparison to the converged full distillation results of \textit{Centipede}, we trained the Centipede agent to 8000 epochs, giving it the same approximate number of update steps as in the fully distilled experiment. Those results are reported, yet note that the reward appeared to plateau at $1000$ epochs. To provide a baseline for our RL agents themselves, we ensured our baseline PPO agents reached comparable performance to the results of DQN agents in the work of \citet{human_level_control}. 

Due to the increase in complexity in Atari environments versus cart-pole environments, the cost of distillation increases greatly as well. It may be unreasonable to commit the large amount of required resources for a full distillation of an Atari environment without knowing how long distillation will take or if distillation can succeed given the experimental setup; thus, we provide a method for controlling the difficulty of distillation by providing intermediate steps between direct-task learning and distillation. We do so by providing an algorithm that generalizes both direct-task learning and distillation. Then, we demonstrate this method on four Atari environments. We perform these experiments by distilling with $k=k_{min}$ for each of the four chosen Atari environments.

\subsection{Encoder Rollback for Variable Difficulty} \label{sec:rollback}

Due to the expense of distilling Atari games, we provide a method to scale the complexity of distillation training, providing evidence for the feasibility of full distillations. We utilize an encoder network to simplify the complex state spaces without reducing the dimensionality of the synthetic dataset. This formulation with the encoder as all but the final layer is defined and explored by \citet{dd_nfr} as a combination of dataset distillation and neural feature regression. While they use this formulation to train downstream task learners, we expand this method for difficulty scaling.

An encoder network $\varepsilon$ with parameters $\xi$ transforms the learner model's input into a smaller space. We use our standard 5-layer Atari learner architecture $\alpha_a$ and split it along layer $l$, such that the encoder $\varepsilon_\xi^l$ consists of layers $[0,l)$ of $\alpha_a$ and the learner $\lambda_\phi^l$ consists of layers $[l,5)$. The encoder's output is fed directly into the learner, just as if they were still a single network. The difference between learner and encoder lies in training: the learner's parameters are still updated by inner learning on the synthetic dataset, and the learner is resampled from $\Lambda^l$ each outer iteration. The encoder, however, is only initialized at the start of training and is trained through outer meta-learning on $T_0$ alongside the distillation, rather than with the learner.

\begin{algorithm}[t]
  \SetAlgoLined
  \SetKwInOut{Input}{input}
  \SetKwInOut{Output}{output}
  
  \Input{distilled dataset $\{X_d, Y_d\}_\theta$, learner distribution $\Lambda^l$, encoder $\varepsilon_\xi^l$, number of inner iterations $n$, target training task $T_0$ with data instances $X_0$ and loss function $L$}
  \Output{distilled dataset $\{X_d, Y_d\}_\theta$ and trained encoder $\varepsilon_\xi^l$}
  \BlankLine
  \While{$\{X_d, Y_d\}_\theta$ and $\varepsilon_\xi^l$ have not converged} {
    $\lambda_\phi^l$ := Sample($\Lambda^l$)\; \label{TDE:samplelearner}
    $e_{X_d}^l$ := $\varepsilon_\xi^l(X_d)$\; \label{TDE:encode_T_d}
    $\phi$, $\nabla_{\phi}$,  := Train($\lambda_\phi^l, e_{T_d}^l$, $Y_d$)\; \label{TDE:train}
    $e^l_{X_0}$ := $\varepsilon_\xi^l(X_0)$\; \label{TDE:encode_T_0}
    $L_{T_0}$ := $L(\lambda_\phi^l, e^l_{X_0})$\; \label{TDE:outerloss}
    $\nabla_\theta$, $\nabla_\xi$ := Backpropagate($\nabla_\phi, L_{T_0}$)\; \label{TDE:backprop}
    Optimize $\theta$ and $\xi$ w.r.t. $\nabla_\theta$ and $\nabla_\xi$\; \label{TDE:outeroptimize}
  }
  return $\{X_d, Y_d\}_\theta$\, $\varepsilon_\xi^l$;
  \caption{Meta-gradient Task Distillation with Encoding($\{X_d, Y_d\}_\theta$, $\Lambda^l$, $\varepsilon_\xi^l$, $n$, $T_0$)}
  \label{alg:TD_Encoding}
\end{algorithm}

We provide the encoder formulation of generalized task distillation as Algorithm \ref{alg:TD_Encoding}. The encoder $\varepsilon$ is initialized with the distilled dataset and, unlike the learner, is not reinitialized in the outer loop. $\varepsilon$ encodes the data associated with both the synthetic task (line \ref{TDE:encode_T_d}) and the real task (line \ref{TDE:encode_T_0}), which in RL-to-SL distillation includes the synthetic data instances and the environment's states, respectively. The learner predicts on the encoded data, rather than directly on the tasks' data (lines \ref{TDE:train}, \ref{TDE:outerloss}). While the learner's parameters are updated through inner learning, the encoder's parameters are updated through outer learning, alongside the distilled dataset (line \ref{TDE:outeroptimize}). We utilize the encoder in inner learning, as that allows the distilled instances to be the same dimensionality throughout our encoder rollback experiments; thus, we can ensure that the distilled dataset is large enough to hold all information needed at any value of $l$.

While distilling with an encoder limits the use of the distillation to training only new learners that utilize that encoder, it does allow scaling the difficulty of distillation for tasks where a full distillation may be too computationally expensive. When $l=0$, the encoder has 0 layers, and thus training is equivalent to distillation without using an encoder, which we refer to as \textit{full distillation}. When $l=5$ (all layers of $\alpha_a$), the learner has 0 layers, and thus the algorithm is equivalent to directly training the encoder network on $T_0$, i.e. direct-task learning. Since direct-task learning is generally more stable and computationally cheaper, higher $l$ values (larger encoders) represent easier distillation tasks, while lower $l$ values (smaller encoders) represent harder distillation tasks. Thus, if we can successfully distill a task with a given encoder, and show that we can roll the encoder back by reducing $l$ and successfully distill the task, we can show how the cost of distillation increases as $l$ decreases. With this, we provide evidence for the feasibility of performing full distillations for tasks for which the cost of distillation is prohibitive given fixed resources.

\subsection{Experiments} \label{sec:atari_experiments}

In the Atari experiments, a 5-layer agent (with architecture $\alpha_a$) is used. These agents are standard for PPO learning and are capable of learning the tested environments directly. We demonstrate that the environments can also be learned with an encoder with $l=4$, where the learner consists of a single linear probe, and for decreasing values of $l$, at the cost of an increase in computational complexity. We test 100 agents on 100 episodes for each distillation experiment.

In general, as $l$ decreases, the number of outer epochs to convergence increases, though the amount by which it increases varies across the environments and network layers. This makes it difficult to accurately predict the cost of a full distillation; however, this method does provide a smoother increase in computation cost per step than jumping from reinforcement learning to distillation. We were able to fully distill \textit{Centipede}, due to its relatively cheap cost of full distillation. The other environments quickly became prohibitively expensive for our resources; however, we assert that full distillations of these environments is possible given sufficient resources  (see Table \ref{tab:atari_kmin}). 

\begin{table*}[tb]
\caption{Results of distillation training with $k=k_{min}$, rolling back an encoder from $l=5$ to $l=0$. $\dagger$ indicates that the run was performed with parallelization (6 gpus), so the number of epochs is not comparable to non-parallelized runs. $*$ indicates the distillation has not fully converged. Empty cells represent experiments that did not exceed random performance due to high computational costs.}
\label{tab:atari_kmin}\vspace{12pt}

\begin{subtable}{\textwidth}
\centering
\caption{Average End-Episode Reward Achieved at Convergence}
\label{subfig:reward_kmin}
\setlength\tabcolsep{3pt}
\begin{tabular}{| c | c | c | c | c | c | c |}
  \hline
   & RL & $l=4$ & $l=3$ & $l=2$ & $l=1$ & Full Distillation \\
  \hline
  
  Centipede & $8011 \pm 1049$ & $8438 \pm 758$ & $7860 \pm 756$ & $8266 \pm 561$ & $8144 \pm 614$ & $8042 \pm 492$ \\
  \hline
  
  Space Invaders & $1508 \pm 320$ & $2123 \pm 156$ & $1443^* \pm 245$ & $768^{*\dagger} \pm 63$ &   & $284^* \pm 2.1$ \\
  \hline

  Ms. Pac-Man & $4310 \pm 521$ & $3351 \pm 91$ & $2950^\dagger \pm 269$ & $2570^\dagger \pm 243$ &   & $415^* \pm 119$ \\
  \hline
  
  Pong & $19.7 \pm 1.2$ & $21.0 \pm 0.01$ & $5.6^{*\dagger} \pm 6.4$ & $-9.1^{*\dagger} \pm 4.1$  &   &   \\
  \hline
\end{tabular}\vspace{12 pt}

\end{subtable}
\begin{subtable}{\textwidth}
\centering
\caption{Outer Epochs to Apparent Convergence}
\label{subfig:epochs_kmin}

\begin{tabular}{| c | c | c | c | c | c | c |}
  \hline
    & RL & $l=4$ & $l=3$ & $l=2$ & $l=1$ & Full Distillation \\
  \hline
  Centipede & 1000 & 3000 & 4000 & 5000 & 40,000 & 8000 \\
  \hline
  Space Invaders & 115,000 & 90,000 &  175,000$^*$ & 670,000$^{*\dagger}$ &   & 205,000$^*$ \\
  \hline
  Ms. Pac-Man & 53,000 & 50,000 &  140,000$^\dagger$ & 1,300,000$^\dagger$ &   & 190,000$^*$ \\
  \hline
  Pong & 5000 & 17,000 & 50,000$^{*\dagger}$ & 300,000$^{*\dagger}$  &   &    \\
  \hline
\end{tabular}\vspace{10pt}
\end{subtable}
\end{table*}
While these experiments provide evidence for the feasibility of full distillations of these environments, the costs may be prohibitive. However, distillation frontloads computation costs and allows for cheaper training on downstream tasks---as low as one optimization step per model. Given a large enough search space for neural architecture search, distillation is a cost-effective solution. Training Atari agents is expensive, and each agent must explore the environment on its own. Distilling an Atari environment allows for training these agents in a single optimization step without exploration. 

\begin{table}[t]
\centering
\caption{Comparison of the costs of distillation vs direct-task learning on Atari \textit{Centipede} (see Table \ref{tab:atari_kmin} for experiment results). Training was timed by performing computation on one NVIDIA A100 GPU. Note that while the initial distillation (``Distilling $\{X_d, Y_d\}_\theta$") is more costly than direct RL training, training new agents on the distilled dataset (``10-shot on $\{X_d, Y_d\}_\theta$") is significantly cheaper.}
\label{tab:time_results}\vspace{5 pt}
\setlength\tabcolsep{1.5pt}
\begin{tabular}{| c | c | c | c |}
  \hline
  Training & Time per iteration & Total iterations & Total Datapoints \\
  \hline
   RL on $E_0$ & 3.25 s/epoch & 1,000 epochs & 8,000,000 \\
   \hline
   Distilling $\{X_d, Y_d\}_\theta$ & 3.73 s/meta-epoch & 8,000 epochs & 64,000,000 \\
   \hline
   10-shot on $\{X_d, Y_d\}_\theta$ & 0.18 s/batch & 1 batch & 10 \\
  \hline
\end{tabular}
\vspace{5pt}
\end{table}

Our distillation of Atari \textit{Centipede} provides a clear example of distillation's cost saving. On our hardware, standard reinforcement learning on \textit{Centipede} took approximately 54 minutes while producing the distillation took 8 hours 17 minutes (see Table \ref{tab:time_results}). Though distillation incurs a 9.2x upfront cost to RL, the benefits are quickly seen by the costs of training new agents on the distilled dataset. It takes 0.18 seconds to initialize a new model and train it on the 10-instance distilled dataset for 1 SGD step. This cost is negligible compared to RL training. If we produce 10 or more agents, distillation will incur less costs overall, including the initial distillation training, than reinforcement learning. If applied to use cases that require large numbers of models, such as ensembling or neural architecture search, the cost benefits are clear, as the costs increase by 0.18 seconds per model, rather than 54 minutes per model through standard RL.

\section{Continuous Environments}

Due to its importance in fields such as robotics, we distill continuous RL environments. We use the MuJoCo-v4 (Multi-Joint dynamics in Contact) environments \citep{mujoco}, which have continuous state and action spaces. Continuous PPO uses the state to predict the mean value of each action and learns a fixed log standard deviation for each action, independent of the state. The distilled dataset is treated similarly to the discrete environments. The distilled state trains the learner's mean action, while an action-agnostic standard deviation is learned. In addition, due to the continuous action space, we opted to distill to $k=64$ instances, rather than determining $k_{min}$ through trial-and-error. We trained all models for 24 hours on 4 GPUs. Note that some models, such as the Inverted Double Pendulum and Walker2D may not have converged. We report our results in Table \ref{tab:mujoco}.

The distillation-trained models outperform the random agents in each environment, and outperform the RL agents in 3 of the 8 environments. This demonstrates that our distillation algorithm is general-use, as it can succeed on both discrete and continuous action spaces. Success on this task specifically opens up the potential for distillation's use in robotics and other real-world RL tasks. Interestingly, the standard deviation of models trained by separate distillations is generally lower than that of the separate RL-trained models, even when the mean reward of the distillation-trained models is higher. This may imply that RL has more outliers, and that the distillation process may favor different minima to those that the RL training finds. Just like in the \textit{Centipede} example, the distilled dataset can train models quickly, in a single step of gradient descent, allowing for many RL agents to be produced at low cost.

\begin{table}[t]
\centering
\caption{Results of 5 distillations on each of the MuJoCo continuous environments. PPO RL performance of 5 agents and uniform random agent performance are also recorded for comparison. For RL, each agent performs 1000 episodes, and for distillation, 1000 episodes are recorded from separate agents trained on the distilled dataset. The means of each RL agent or distilled dataset are taken, and the mean and standard deviation across the 5 agents/datasets are recorded.}
\label{tab:mujoco}\vspace{5pt}
\setlength\tabcolsep{1.5pt}
\begin{tabular}{| c | c | c | c |}
  \hline
   & Random & RL & Full Distillation \\
  \hline
  Ant & $-52 \pm 88$ & $3980 \pm 669$ & $3115 \pm 319$ \\
  \hline
  Half Cheetah & $-282 \pm 79$ & $1908 \pm 1347$ & $2226 \pm 242$ \\
  \hline
  Hopper & $21 \pm 23$ & $1441 \pm 532$ & $2168 \pm 302$  \\
  \hline
  Humanoid & $123 \pm 37$ & $631 \pm 486$ & $544 \pm 32$  \\
  \hline
  H. Standup & $34,034 \pm 3028$ & $104,950 \pm 53,667$ & $138,478 \pm 6037$ \\
  \hline
  Inv. Double Pend. & $57 \pm 16$& $7553 \pm 1036$ & $1235 \pm 427$ \\
  \hline
  Reacher & $-42.95 \pm 4.05$ & $-5.63 \pm 0.22$ & $-7.78 \pm 0.57$ \\
  \hline
  Walker2D & $2.1 \pm 5.1$ & $3537 \pm 432$ & $331.7 \pm 117$ \\
  \hline
\end{tabular}
\vspace{5pt}
\end{table}

\section{Discussion}

We have demonstrated distillation of cart-pole, Atari, and MuJoCo environments into compact, single-batch datasets. We have demonstrated the benefits of distillation in reducing training costs. While the up-front expenses of distillation are not insignificant, for applications that require training many models, the per-model speedup is significant. The cost may be reduced by using better reinforcement learning algorithms, better hyperparameters, and parallelization. The performance of distillation-trained models may be improved with a longer distillation process or a robust hyperparameter search.

Distillation has been proposed for use in downstream tasks such as data anonymity \citep{data_anonymity} and neural architecture search \citep{dataset_distillation},
and could be viable for other search problems, such as finding high-performance models or good parameter initializations. In addition, methods that require multiple end-task models, such as ensembling, may benefit from distillation.
However, distillation has not yet been widely adopted for these use cases, likely due to applications being an afterthought in distillation research and to the computational expense of distillation. Generalized task distillation, and RL-to-SL distillation in particular, opens avenues for real-world application by expanding the types of problems that distillation can be used on.

We intend for our future work to examine use cases of task distillation to prove its viability for real-world applications, such as neural architecture search. We hope to improve interpretability and understanding of how and why distillation works. We also intend to expand distillation to other real and synthetic tasks, such as more complex RL algorithms.

\bibliography{main}

\newpage
\newpage

\appendix
\section{Appendix - Hyperparameters and Additional Details}

\begin{table}[ht]
    \centering
    \begin{tabular}{| c | c |}
        \hline
        Distillation Hyperparameters & Values \\
        \hline
        distiller optimizer & Adam \\
        \hline
        critic optimizer & Adam \\
        \hline
        inner optimizer & SGD \\
        \hline
        distiller lr & $2.5*10^{-4}$ \\
        \hline
        critic lr & $2.5*10^{-4}$ \\
        \hline
        initial inner lr & $2 * 10^{-2}$ \\
        \hline
        outer objective & meta-PPO \\
        \hline
        inner objective & mean squared error \\
        \hline
        episodes per epoch & 10\\
        \hline
        reward discount $\gamma$ & 0.99 \\
        \hline
        policy epochs & 4 \\
        \hline
        outer batch size & 512 \\
        \hline
        rollout length & 200 \\
        \hline
    \end{tabular}
    \caption{Hyperparameters utilized in experiments.}
    \label{tab:hyperparameters}
\end{table}

\begin{table}[ht]
    \centering
    \begin{tabular}{| c | c | c |}
        \hline
        Environment & Number of Actions & $k_{min}$ \\
        \hline
        Centipede & 18 & 10\\
        \hline
        Ms. Pac-Man & 9 & 6\\
        \hline
        Pong & 6 & 4\\
        \hline
        Space Invaders & 6 & 4\\
        \hline
    \end{tabular}
    \caption{Action spaces of Atari environments used.}
    \label{tab:action_spaces}
\end{table}

\section{ND Cart-pole} \label{apx:ndcp}

In an effort to create a reinforcement environment with a controllable state and action space, we have expanded the classic control cart-pole problem from a 1 degree-of-freedom problem to an arbitrary $N$ degree-of-freedom problem. This problem contains $N$ dimensions in which the cart can be moved and in which the pole can fall. The observation state consists of $N$ sets of quadruples: the cart's position, the cart's velocity, the pole's angle, and the pole's angular velocity in the corresponding dimension; or as 4 vectors of size $N$. We provide a diagram showing two views of 2D cart-pole in Figure \ref{fig:ndcp}

At each timestep, a force of magnitude $F$ is applied to the cart in a direction selected by the agent. To limit the action space to $2N$, we limit the applied force to $\pm F$ along one of the $N$ axes, resulting in a force of $0$ along all other axes (other than gravity pulling the pole downward). This is similar to the classic 1D cart-pole problem, where the agent can select to apply a force of $\pm F$ to the cart along the dimension of the track. Another way of framing ND cart-pole is as $N$ independent cart-pole environments, in which the agent applies force $\pm F$ in exactly one environment per timestep, and the other environments receive a force of $0$. If the agent fails in any of the parallel cart-poles, it fails in all. Like in standard cart-pole, each iteration provides a reward of +1 until the agent fails. In all ND cart-pole experiments, we terminate the environment once an accumulated reward of 500 is reached, an arbitrary reward ceiling often used in standard cart-pole (as unbounded cart-pole can lead to extremely long episodes).

This environment allows for easy difficulty scaling for reinforcement learning. However, the scaling does not appear to be linear. With a sufficiently high $N$, the agent cannot act in all dimensions before the pole has a chance to fall in one of them. This can be counteracted by altering the time scale (i.e. scaling down velocity and acceleration), giving the agent a quicker reaction time. In our experiments, however, we only utilize up to 5D cart-pole, which can be solved by our PPO agent without changing the time scale.

\begin{figure*}[ht]
        \centering
        \includegraphics[width=\textwidth]{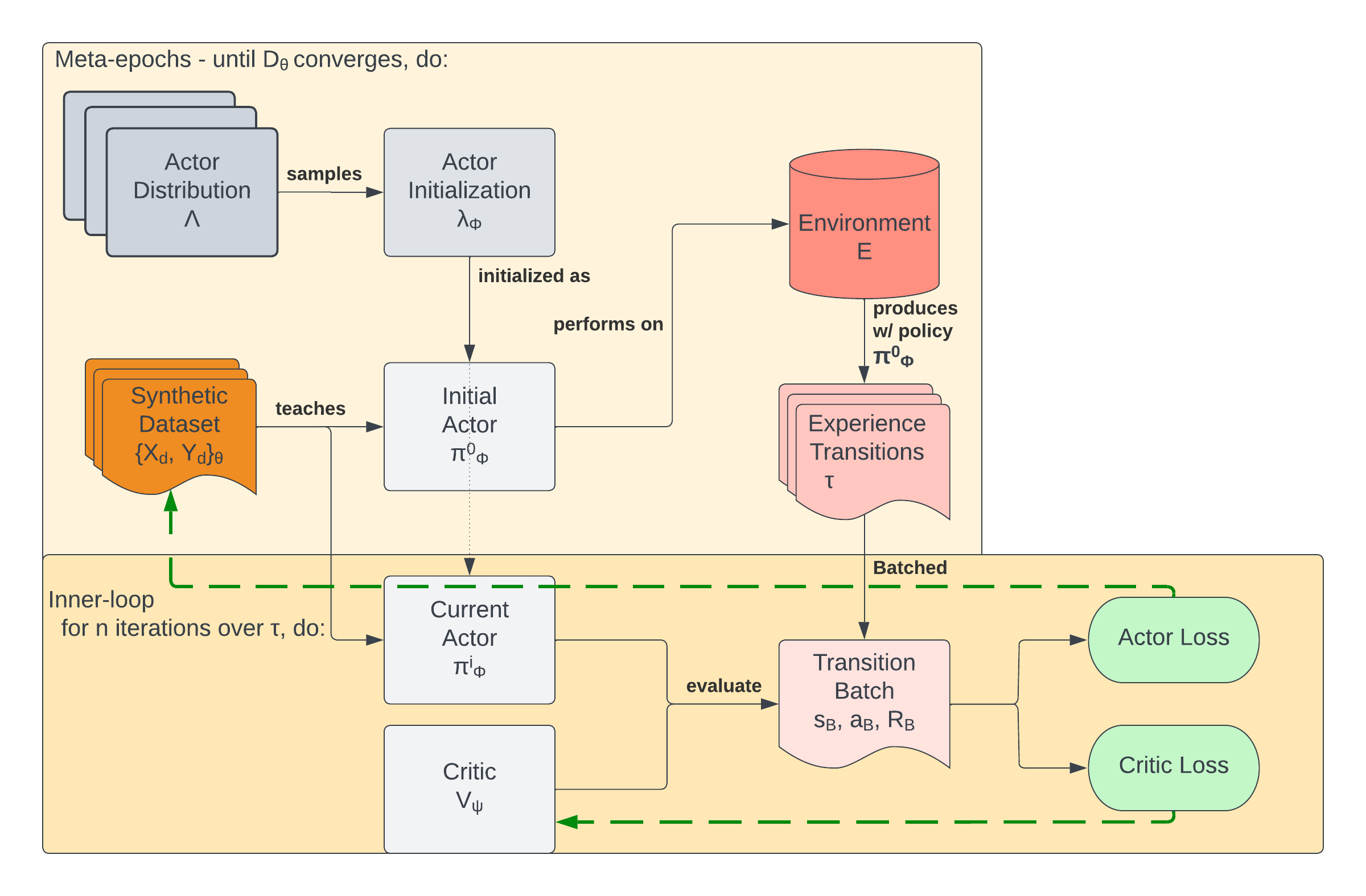}
        \caption{Training process for RL-to-SL distillation with generic actor-critic method.}
        \label{fig:PPODistill}
\vspace{15pt}
\end{figure*}

\begin{figure*}[ht]
\centerline{\includegraphics[width=.60\textwidth]{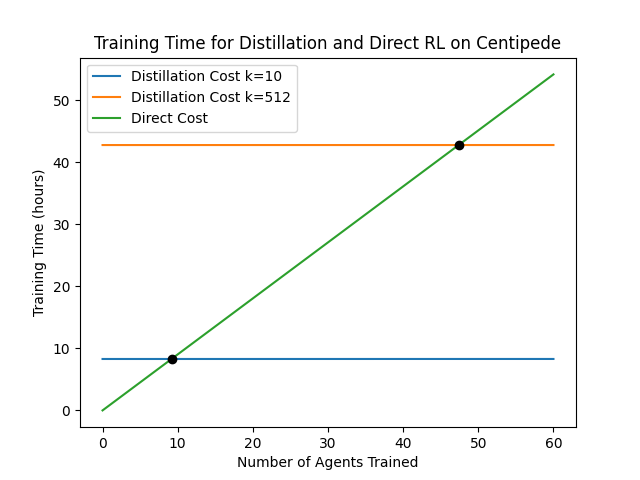}}
\caption{Comparing \textit{Centipede} training time based on number of agents trained between direct-task learning (reinforcement learning), distillation with $k=512$, and distillation with $k=10$ ($k_{min}=10$ for \textit{Centipede}). Note that both distillation lines are parallel with a slope of $0.18$ seconds (the cost of training each learner with the final distilled set). Direct-task learning is cheaper than distillation training; however, training each agent after distillation using the distilled set is significantly cheaper than direct-task learning. Distillation training with $k=10$ is significantly cheaper than distillation training with $k=512$.} 
\label{fig:distill_vs_direct_learning}
\end{figure*}

\begin{table*}[ht]
    \begin{subtable}{.45\textwidth}
    \centering
        \begin{tabular}{|c | c | c | c | c |}
        \hline
         & h=32 & h=64 & h=128 & h=256\\
        \hline
        L=0 & 500.0 & 500.0 & 500.0 & 500.0\\
        \hline
        L=1 & 500.0 & 500.0 & 500.0 & 500.0\\
        \hline
        L=2 & 500.0 & 500.0 & 500.0 & 500.0\\
        \hline
        L=3 & 500.0 & 500.0 & 500.0 & 500.0\\
        \hline
        L=4 & 499.9 & 500.0 & 500.0 & 500.0\\
        \hline
        L=5 & 494.9 & 499.5 & 500.0 & 500.0\\
        \hline
        \end{tabular}
        \caption{2-shot 1D}
    \end{subtable}
    \vspace{12pt}
    \hfill
    \begin{subtable}{.45\textwidth}
    \centering
        \begin{tabular}{|c | c | c | c | c |}
        \hline
         & h=32 & h=64 & h=128 & h=256\\
        \hline
        L=0 & 94.4 & 90.4 & 93.1 & 93.2\\
        \hline
        L=1 & 490.5 & 490.5 & 491.8 & 491.8\\
        \hline
        L=2 & 500.0 & 500.0 & 500.0 & 500.0\\
        \hline
        L=3 & 492.7 & 500.0 & 500.0 & 500.0\\
        \hline
        L=4 & 447.8 & 490.4 & 499.2 & 500.0\\
        \hline
        L=5 & 371.3 & 448.3 & 489.4 & 499.6\\
        \hline
        \end{tabular}
        \caption{3-shot 2D}
    \end{subtable}

    \begin{subtable}{.45\textwidth}
    \centering
        \begin{tabular}{|c | c | c | c | c |}
        \hline
        L=0 & 78.7 & 76.9 & 78.6 & 80.4\\
        \hline
        L=1 & 430.7 & 468.7 & 482.7 & 488.2\\
        \hline
        L=2 & 382.8 & 447.6 & 486.6 & 498.2\\
        \hline
        L=3 & 292.7 & 368.4 & 441.5 & 488.1\\
        \hline
        L=4 & 220.7 & 284.8 & 366.0 & 437.1\\
        \hline
        L=5 & 170.7 & 221.8 & 287.0 & 374.9\\
        \hline
        \end{tabular}
        \caption{4-shot 3D}
    \end{subtable}
    \vspace{12pt}
    \hfill
    \begin{subtable}{.45\textwidth}
    \centering
        \begin{tabular}{|c | c | c | c | c |}
        \hline
        L=0 & 49.1 & 50.7 & 51.1 & 50.9\\
        \hline
        L=1 & 177.7 & 193.1 & 200.6 & 208.0\\
        \hline
        L=2 & 182.0 & 210.2 & 235.7 & 261.0\\
        \hline
        L=3 & 159.1 & 186.5 & 217.8 & 251.3\\
        \hline
        L=4 & 135.6 & 162.1 & 189.6 & 225.2\\
        \hline
        L=5 & 109.5 & 136.8 & 161.9 & 198.8\\
        \hline
        \end{tabular}
        \caption{5-shot 4D}
    \end{subtable}

    \begin{subtable}{.45\textwidth}
    \centering
        \begin{tabular}{|c | c | c | c | c |}
        \hline
        L=0 & 38.0 & 39.6 & 39.1 & 39.4\\
        \hline
        L=1 & 125.0 & 134.5 & 140.0 & 143.4\\
        \hline
        L=2 & 149.3 & 165.5 & 183.2 & 196.8\\
        \hline
        L=3 & 135.1 & 153.8 & 175.1 & 197.5\\
        \hline
        L=4 & 113.0 & 136.3 & 157.2 & 181.5\\
        \hline
        L=5 & 79.5 & 115.1 & 136.0 & 160.9\\
        \hline
        \end{tabular}
        \caption{6-shot 5D}
    \end{subtable}

    \vspace{12pt}

    \caption{Extended results for Table 1 in the main paper, obtained by performing k-shot learning and validating on ND cart-pole, using a grid search over L ( $[0,5]$ hidden layers) and h ($2^{[5,8]}$ hidden layer width). Error bars are omitted for brevity, but were of similar magnitude to those of Table 1 in the main paper.}
    \label{tab:lh}
\end{table*}

\begin{table*}
\begin{subtable}{\textwidth}
\centering
\begin{tabular}{| c | c | c | c | c | c | c |}
  \hline
   & RL & $l=4$ & $l=3$ & $l=2$ & $l=1$ & Full Distillation \\
  \hline
  Centipede & 8378 & 10,955 & 8251 & 7874 & 8327 & 7694 \\
  \hline
  Space Invaders & 1947 & 2941 & 1732 &   &  & 277$^*$ \\
  \hline
  Ms. Pac-Man & 3894 & 2987 & 731$^*$ & 699$^*$ & 342$^*$ & 518$^*$ \\
  \hline
  Pong & 17.3 & 20.5 &   &   &   &   \\
  \hline
\end{tabular}
\caption{Average End-Episode Reward Achieved at Convergence}
\label{subfig:reward}
\end{subtable}

\vspace{12pt}

\begin{subtable}{\textwidth}
\centering
\setlength\tabcolsep{2pt}
\begin{tabular}{| c | c | c | c | c | c | c |}
  \hline
    & RL & $l=4$ & $l=3$ & $l=2$ & $l=1$ & Full Distillation \\
  \hline
  Centipede & 1,000 & 1,500 & 3,000 & 2,000 & 55,000 & 40,000 \\
  \hline
  Space Invaders & 115,000 & 195,000 & 250,000 &   &  & 30,000$^*$  \\
  \hline
  Ms. Pac-Man & 53,000 & 110,000 & 55,000$^*$ & 75,000$^*$ & 90,000$^*$ & 85,000$^*$ \\
  \hline
  Pong & 5,000 & 40,000 &   &   &   &   \\
  \hline
\end{tabular}
\caption{Outer Epochs to Apparent Convergence}
\label{subfig:epochs}
\end{subtable}

\vspace{12pt}

\caption{
Results for $k=512$ distillation, compare to Table \ref{tab:atari_kmin}.
}
\label{tab:atari}
\end{table*}

\begin{table*}
\begin{subtable}{\textwidth}
\centering
\begin{tabular}{|c|c|c|c|c|}
\hline
1D & 2D & 3D & 4D & 5D \\
\hline
$21.8 \pm 11.6$ & $18.9 \pm 6.6$ & $18.5 \pm 4.9$ & $18.6 \pm 4.3$ & $18.6 \pm 3.8$ \\
\hline
\end{tabular}
\caption{ND cart-pole}
\end{subtable}

\vspace{12pt}

\begin{subtable}{\textwidth}
\centering
\begin{tabular}{|c|c|c|c|c|}
\hline
Centipede & Ms. PacMan & Space Invaders & Pong \\
\hline
$2443 \pm 1288$ & $239.6 \pm 80.3$ & $165.6 \pm 124.6$ & $-20.3 \pm 1.0$ \\
\hline
\end{tabular}
\caption{Atari}
\end{subtable}

\vspace{12pt}

\label{tab:cp_random}
\caption{Performance of uniform random agents.}
\end{table*}

\begin{figure*}[ht]
    \centering
    \begin{subfigure}[b]{0.32\textwidth}
        \centering
        \includegraphics[width=\textwidth]{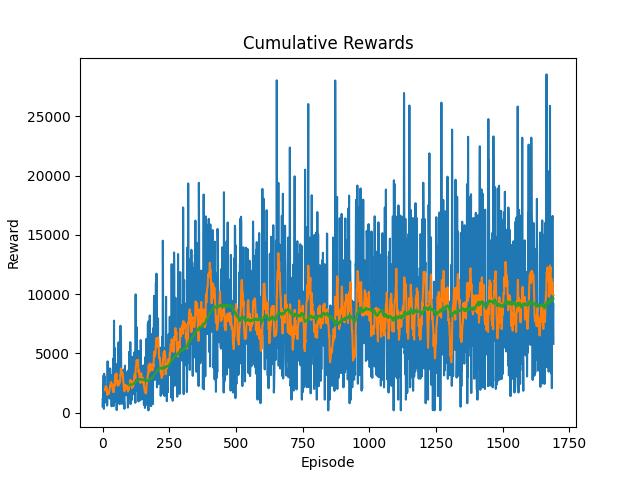}
        \caption{$l$=5 (RL)}
        \label{sfig:ckmin_rl}
    \end{subfigure}
    \hfill
    \begin{subfigure}[b]{0.32\textwidth}
        \centering
        \includegraphics[width=\textwidth]{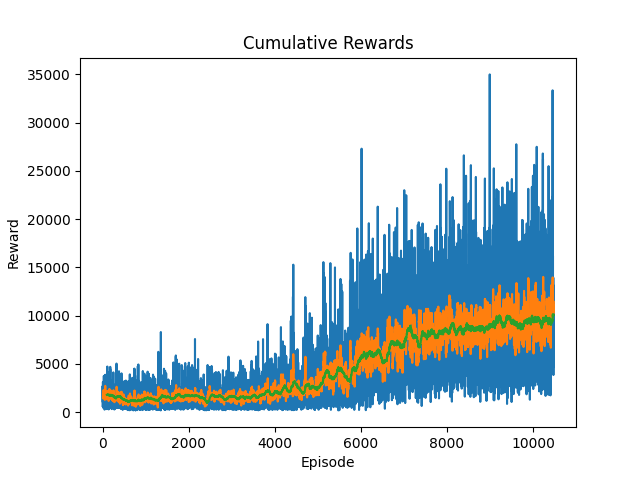}
        \caption{$l$=4}
        \label{sfig:ckmin_l4}
    \end{subfigure}
    \hfill
    \begin{subfigure}[b]{0.32\textwidth}
        \centering
        \includegraphics[width=\textwidth]{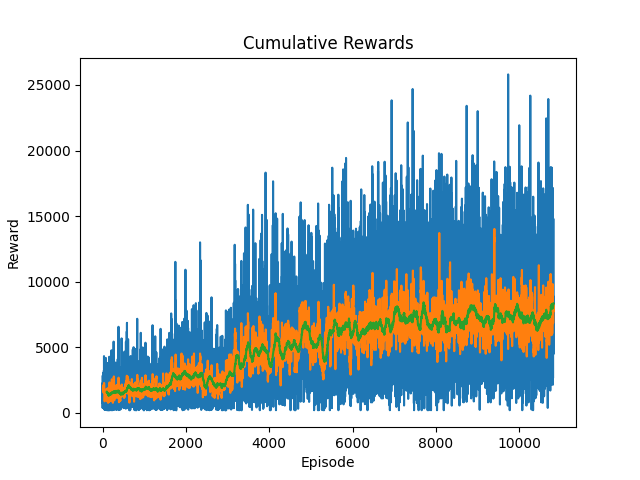}
        \caption{$l$=3}
        \label{sfig:ckmin_l3}
    \end{subfigure}

    \vspace{12pt}

    \begin{subfigure}[b]{0.32\textwidth}
        \centering
        \includegraphics[width=\textwidth]{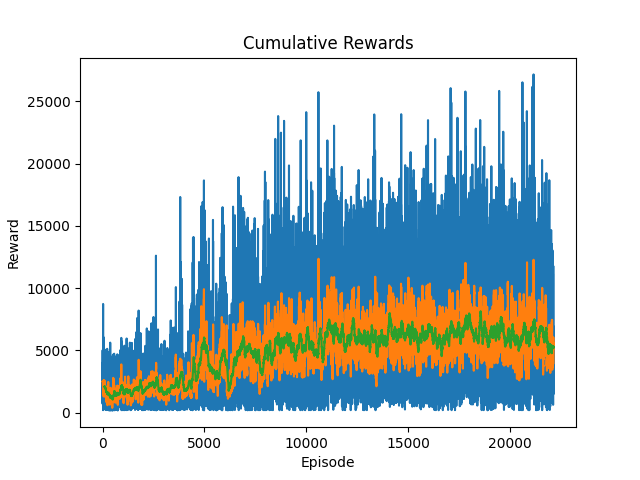}
        \caption{$l$=2}
        \label{sfig:ckmin_l2}
    \end{subfigure}
    \hfill
    \begin{subfigure}[b]{0.32\textwidth}
        \centering
        \includegraphics[width=\textwidth]{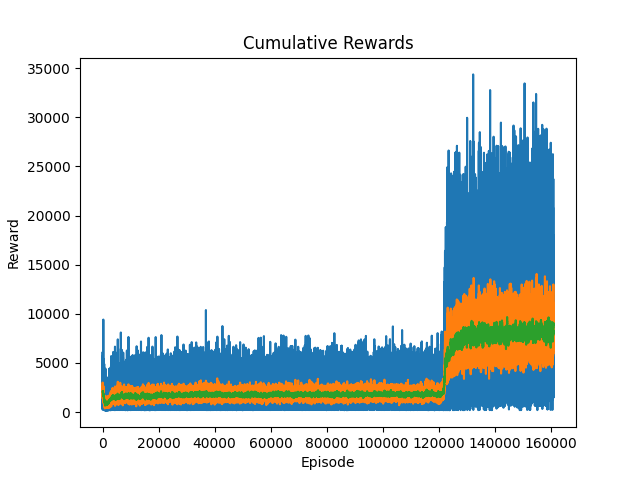}
        \caption{$l$=1}
        \label{sfig:ckmin_l1}
    \end{subfigure}
    \hfill
    \begin{subfigure}[b]{0.32\textwidth}
        \centering
        \includegraphics[width=\textwidth]{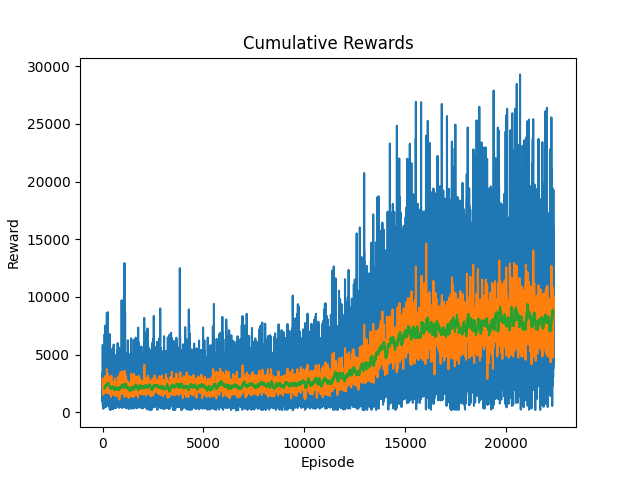}
        \caption{$l$=0 (Full Distillation)}
        \label{sfig:ckmin_distill}
    \end{subfigure}

    \vspace{12pt}
    
    \caption{End-episode rewards for outer learning on \textit{Centipede} distillation with $k=10$ ($k_{min}$).}
    \label{fig:centipede_rewards_kmin}
\end{figure*}

\newpage

\section{Diffusion as a Distillation Baseline}

In order to provide a baseline for the distillation process, we trained a diffusion model to learn the state-action distribution of an expert Centipede RL agent, and used the data generated from the diffusion model to train RL agents.

We trained a single diffusion model until convergence on a dataset of 50,000 state-action pairs from the policy of the RL agent reported in our results.

We trained RL models on the diffusion-generated dataset. First, we trained an RL agent on 1600 generated pairs (taking approximately 19 hours, approximately doubling the time requirements of distillation training). This agent was tested over 1000 episodes on Centipede and got an average reward of 2067.

We then tested a cheaper method of generating a single-batch dataset of 32 instances (2 per action class) and allowing the models to train until convergence. This more than triples the data usage of distillation. We tested this on 5 agents over 1000 episodes each and got an overall mean reward of 1708; the best performing model got a mean reward of 1980.

The Centipede distillation, with a mean reward of 8083 far surpassed these baselines. It beat the first in compression size and time, including distillation training. The second method was much faster to train than to train a distillation, but the end-training of distillation is significantly faster and uses less data. These results demonstrate that distillation training is not easily replaceable with generative AI methods trained on a near-optimal state-action distribution.

\section{Additional Figures and Results} \label{apx:figures}

Here we provide figures and tables that did not fit into the main publication.

\end{document}